\documentclass[11pt]{article}

\usepackage[final]{acl}

\usepackage{times}
\usepackage{latexsym}
\usepackage{algorithm}      
\usepackage{algpseudocode}  
\usepackage{amsmath}        
\usepackage{hyperref}
\usepackage{multirow}
\usepackage[T1]{fontenc}
\usepackage{tabularx}
\usepackage{booktabs}
\usepackage{enumitem}
\usepackage[utf8]{inputenc}

\usepackage{microtype}

\usepackage{inconsolata}

\usepackage{graphicx}

\title{Rewarding Intellectual Humility: Learning When Not to Answer in Large Language Models}

\author{
  \textbf{Abha Jha}, \textbf{Akanksha Mahajan}, \textbf{Ashwath Vaithinathan Aravindan} \\
  \textbf{Praveen Saravanan}, \textbf{Sai Sailaja Policharla},          \textbf{Sonal Chaturbhuj Gehlot} \\
  University of Southern California \\
  Los Angeles, United States of America \\
  \texttt{\{abhajha,am59710,vaithina,psaravan,polichar,sgehlot\}@usc.edu}
}

\begin{document}
\maketitle
\renewcommand{\abstractname}{Abstract}
\begin{abstract}
Large Language Models (LLMs) often produce hallucinated or unverifiable content, undermining their reliability in factual domains. This work investigates Reinforcement Learning with Verifiable Rewards (RLVR) as a training paradigm that explicitly rewards abstention (“I don’t know”) alongside correctness to promote intellectual humility. We fine-tune and evaluate Granite-3.3-2B-Instruct and Qwen-3-4B-Instruct on the MedMCQA and Hendrycks Math benchmarks using a ternary reward structure 
($-1$, r\textunderscore abs, 1) under varying abstention reward structures. We further study the effect of combining RLVR with supervised fine-tuning strategies that teach abstention prior to reinforcement learning. Our results show that moderate abstention rewards (r\textunderscore abs $\approx -0.25$ to 0.3) consistently reduce incorrect responses without severe accuracy degradation on multiple-choice tasks, with larger models exhibiting greater robustness to abstention incentives. On open-ended question answering, we observe limitations due to insufficient exploration, which can be partially mitigated through supervised abstention training. Overall, these findings demonstrate the feasibility and flexibility of verifiable reward design as a practical approach for hallucination mitigation in language models. Reproducible code for our abstention training framework is available \href{https://github.com/Mystic-Slice/rl-abstention}{here}.
\end{abstract}

\section{Introduction}

In this project, we present progress on Reinforcement Learning with Verifiable Rewards (RLVR) \cite{lambert2025tulu3pushingfrontiers, deepseekai2025deepseekr1incentivizingreasoningcapability} with a modified reward setup as a framework that incentivizes models to exhibit intellectual humility by rewarding correct answers and partial abstention (“I don’t know”), while penalizing incorrect responses. We conducted experiments using Granite-3.3-2B-Instruct \cite{ibmGranite33_2BInstruct} and Qwen-3-4B-Instruct \cite{qwen3_4b_instruct_2507} models on the MedMCQA \cite{medmcqa} and Hendrycks MATH \cite{hendrycksmath2021} dataset. Our findings suggest that moderate abstention rewards can reduce hallucinations without severely impacting overall task accuracy, supporting RLVR as a promising approach for trustworthy language model alignment.

Large Language Models (LLMs) \cite{zhao2025surveylargelanguagemodels} have become ubiquitous across professional and research domains but continue to suffer from hallucinations confidently producing factually incorrect or unverifiable information \cite{Huang_2025}. This issue arises in part because LLMs are never explicitly trained to abstain when uncertain; their objective always rewards response generation. Existing mitigation methods such as Supervised Fine-Tuning (SFT) \cite{rtuning} and Reinforcement Learning from Human Feedback (RLHF) \cite{rlhf_abstention} reduce hallucinations but rely heavily on costly, human-in-the-loop feedback. We hypothesize that introducing an explicit abstention reward $r_{abs}$ will reduce incorrect responses (hallucinations) while allowing a tunable trade-off with accuracy.

In this work, we present a modified Reinforcement Learning with Verifiable Rewards (RLVR) framework that explicitly rewards explicit abstention (“I don't know”) in addition to rewarding correct answers and penalizing incorrect ones. We evaluate this ternary reward scheme on multiple models (Granite-3.3-2B-Instruct and Qwen-3-4B-Instruct) and two verifiable QA datasets (MedMCQA and Hendrycks Math). Our experiments show that appropriately chosen abstention rewards ($r_{abs}$) reduce the rate of incorrect (hallucinatory) outputs while providing a tunable trade-off with accuracy; modest positive $r_{abs}$ values can substantially reduce incorrect answers with limited loss in correct answers on MCQ datasets.






\section{Related work}

Alignment methods such as supervised fine-tuning (SFT) and reinforcement learning from human feedback (RLHF) have been shown to reduce hallucinations but not eliminate them. \cite{ouyang2022training} introduced SFT followed by RLHF using human demonstrations, while \cite{bai2022training} optimized preference models to balance helpfulness and harmlessness; both approaches rely on costly human feedback and do not explicitly model uncertainty. \cite{lambert2025tulu3pushingfrontiers} surveyed these methods, identifying hallucination as a persistent failure mode and motivating the need for more verifiable reward structures.

More closely related, \cite{godin2019learning} proposed a ternary reward structure that explicitly rewards abstention in question answering, improving precision and reducing hallucinations. However, this work predates modern instruction-tuned LLMs and does not study abstention within contemporary SFT or RL-based training pipelines. In contrast, our work integrates abstention-aware supervision into modern LLM alignment and evaluation, enabling calibrated refusal behavior without relying solely on dense human preference feedback. 

\section{Methodology and Experiment Setup}
\label{Section:3}

\subsection{Datasets}
We conduct our experiments on two verifiable question-answering datasets. MedMCQA ~\cite{medmcqa} [\href{https://huggingface.co/datasets/openlifescienceai/medmcqa}{Dataset Link}] is a dataset with multiple-choice questions from real-world medical entrance exams. It contains 4 options including the identified ground truth. We augment the options with an additional "I Don't Know" option to evaluate the model's ability to abstain when it could not answer correctly. We also additionally test our technique on Hendrycks Math \cite{hendrycks2021measuring}[\href{https://huggingface.co/datasets/EleutherAI/hendrycks_math}{Dataset Link}], a dataset with hard math questions, most of which require proof over multiple steps to answer the question. This dataset is to test the generalization of our technique to open-ended question answering, which is the ultimate goal of the project.

\subsection{Model Selection}
We evaluate our approach on two instruction-tuned LLMs spanning multiple scales and architectures.
Granite-3.3-2B-Instruct (IBM) serves as a compact baseline for controlled ablation, and Qwen-3-4B-Instruct \cite{qwen3} acts as a high-capacity multilingual model known for occasional hallucinations.

\subsection{Prompt Design}
\label{sec:prompt}
To ensure compatibility with verifiable reward computation and to explicitly enable abstention behavior, we design task-specific prompts for each dataset. In all cases, the prompts explicitly instruct the model to answer only when confident and to otherwise respond with an abstention token (``I don't know'').

\paragraph{MedMCQA:} For the multiple-choice MedMCQA dataset, we enforce a structured response format to facilitate automatic verification and reward computation.
\begin{quote}
\small
\texttt{Answer the following question. Provide your thoughts between <reasoning> and </reasoning> symbols. Provide the final answer option (letter only) between <answer> and </answer> symbols. Answer only if you are certain, else choose I Don't Know.}
\end{quote}

The strict use of XML-style tags enables regex-based validation of output structure during training, while the explicit abstention instruction ensures that ``I Don't Know'' is treated as a valid and verifiable action under the ternary reward scheme.

\paragraph{Hendrycks Math:} For the open-ended mathematical reasoning tasks, we use a minimal prompt that encourages step-by-step reasoning while constraining the final answer format for verifiability.
\begin{quote}
\small
\texttt{Answer the following question. Provide your reasoning. Then, provide the final answer in \textbackslash boxed\{YOUR\_ANSWER\_HERE\} format. Answer only if you are certain, else output \textbackslash boxed\{I Don't Know\}.}
\end{quote}

\subsection{Reward Design}
Following the idea of ternary reward structures for abstention in QA \cite{godin2019learning}, we extend Reinforcement Learning with Verifiable Rewards (RLVR) \cite{lambert2025tulu3pushingfrontiers, deepseekai2025deepseekr1incentivizingreasoningcapability} to explicitly encourage intellectual humility.
Each model output must conform to a verifiable schema defined as per equation \ref{eq:1}, where y is model's answer:
\begin{equation}
R(y) =
\begin{cases}
    r_{\text{correct}}, & \text{if $y$ is a correct answer} \\
    r_{\text{abs}}, & \text{if $y =$ ``I don't know''} \\
    r_{\text{wrong}}, & \text{if $y$ is an incorrect answer}  
\end{cases}
\label{eq:1}
\end{equation}
with $r_{correct} = 1$, $r_{wrong} = -1$, and $r_{abs}$ as a hyperparameter.
This is the accuracy reward.

Additionally, with MedMCQA, a format reward verifies syntactic compliance using a regex constraint that enforces \texttt{<reasoning>}…\texttt{</reasoning>} \texttt{<answer>}…\texttt{</answer>} tags in the model output. Such a strict format is not enforced when training on the Hendrycks Math in order to preserve the open-ended answering ability.

This formulation aligns with verifiable-reward learning in reasoning-based LLMs \cite{deepseekmath} and minimizes dependence on human preference models typical of RLHF \cite{bai2022training, rlhf_abstention}.

\subsection{Training Framework and Baselines}
We build on the Group Relative Policy Optimization (GRPO) algorithm \cite{deepseekmath}, integrated through the Hugging Face Transformer Reinforcement Learning (TRL) library.

Our experiments comprise three settings:
\begin{enumerate}[noitemsep, topsep=0pt]
    \item \textbf{RL-only:} models trained directly with the ternary reward signal \ref{eq:1}.
    \item \textbf{RL-SFT-Random:} models initially SFT trained on the MedMCQA dataset with 30\% of randomly selected ground truths set to "I don't know" to encourage abstention and further fine-tuned with RLVR.
    \item \textbf{RL-RTuning:} models initially SFT trained on the MedMCQA dataset with answers to questions that the base model got wrong set to "I don't know", based on the R-Tuning work\cite{rtuning}, to encourage abstention and further fine-tuned with RLVR.
\end{enumerate}


Baselines include:
\begin{itemize}[noitemsep, topsep=0pt]
    \item \textbf{No-IDK baseline:} standard multiple-choice QA without abstention.
    \item \textbf{IDK-enabled baseline:} five-option (A–E) QA allowing “I don’t know.”
\end{itemize}



\subsection{Training Setup}
We train models using Group Relative Policy Optimization (GRPO) with a ternary accuracy-based reward. Unless stated otherwise, all experiments use LoRA-based fine-tuning.

Experiments are run using the TRL GRPOTrainer with the following configuration:
\begin{itemize}[noitemsep, topsep=0pt]
    \item Optimizer learning rate: $2 \times 10^{-5}$
    \item Gradient accumulation steps: 64
    \item Per-device batch size: 8
    \item Number of generations per prompt: 8
    \item Maximum training steps: 500
    \item Max prompt length: 256 tokens
    \item Max completion length: 1024 tokens
    \item Precision: bfloat16
    \item Gradient checkpointing enabled
\end{itemize}

Training uses vLLM to speed up response generation in colocated mode with GPU memory utilization capped at 0.6. Checkpoints are saved every 10 steps, and evaluation is performed every 20 steps. Experiments are run on NVIDIA A40 and A100 GPUs. Each training run was allowed to proceed for approximately 24 hours of wall-clock time. Models were trained on approximately 8k samples and evaluated on a held-out test set of around 100 samples.

\section{Results}
\label{Section:4}

The models' response distribution for the base models and the ones trained with different RLVR-based training approaches for MedMCQA is shown in Table \ref{tab:granite_qwen_medmcqa}. For each model, the baseline without IDK represents conventional QA behavior, while RLVR variants correspond to different abstention-reward settings ($r_{abs}$). Table \ref{tab:granite_qwen_math} shows the results for the Granite-3.3-2B-Instruct model on the Hendrycks Math data. The two sets of results show that the optimal reward structure varies with different models and datasets.

\begin{table}[h!]
\centering
\caption{Comparison of Granite-3.3-2B-Instruct and Qwen-3-4B-Instruct performance on MedMCQA for different training methods.}
\resizebox{\columnwidth}{!}{%
\begin{tabular}{llccc}
\toprule
\textbf{Model} & \textbf{Method} & \textbf{Correct (\%)} & \textbf{Incorrect (\%)} & \textbf{IDK (\%)} \\
\midrule
\multirow{4}{*}{Granite-3.3-2B-Instruct} 
& Base model & 46.8 & 53.2 & -- \\
& Base model (with IDK option) & 44.8 & 48.6 & 6.6 \\
& RL-only (-0.25) & 46.4 & 46.4 & 7.0 \\
& RL-SFT-Random (-0.25) & 39.0 & \textbf{40.0} & 21.0 \\
& RL-RTuning (-0.25) & 47.0 & 45.0 & 8.0 \\
\midrule
\multirow{3}{*}{Qwen-3-4B-Instruct} 
& Base model & 67.5 & 32.5 & -- \\
& Base model (with IDK option) & 65.4 & 27.3 & 7.3 \\
& RL-only (0.3) & 48.0 & \textbf{10.3} & 41.0 \\
& RL-SFT-Random (0.1) & 51.0 & 28.0 & 21.0 \\
\bottomrule
\end{tabular}%
}
\label{tab:granite_qwen_medmcqa}
\end{table}

\begin{table}[h!]
\centering
\caption{Comparison of Granite-3.3-2B-Instruct performance on Hendrycks Math for different training methods.}
\resizebox{\columnwidth}{!}{%
\begin{tabular}{llccc}
\toprule
\textbf{Model} & \textbf{Method} & \textbf{Correct (\%)} & \textbf{Incorrect (\%)} & \textbf{IDK (\%)} \\
\midrule
\multirow{4}{*}{Granite-3.3-2B-Instruct} 
& Base model & 49.59 & 50.41 & -- \\
& Base model (with IDK option) & 45.99 & 53.99 & 0.03 \\
& RL-only (-0.25) & 66.0 & 34.0 & 0.0 \\
& RL-SFT-Random (-0.5) & 35.0 & \textbf{26.0} & 39.0 \\
& RL-RTuning (-0.5) & 21.3 & 33.7 & 44.4 \\
\bottomrule
\end{tabular}%
}
\label{tab:granite_qwen_math}
\end{table}

\subsection{RL-only: Rewarding abstention}

Table~\ref{tab:rlonly_rewards_combined} summarizes the RL-only experiments for both Granite-3.3-2B-Instruct and Qwen-3-4B-Instruct across a range of abstention rewards on the MedMCQA dataset. 
In both models, moving from negative to positive $r_{abs}$ values consistently increases abstention rates resulting in lowering hallucination but at the cost of accuracy. 
Granite-3.3-2B-Instruct transitions from near-zero abstention at $r_{abs} = -0.5$ to full abstention at $r_{abs} = 0.3$, confirming the model’s high reward sensitivity. 
Qwen-3-4B-Instruct shows a similar monotonic trend but retains higher accuracy across all settings, with $r_{abs} = 0.3$ yielding 41.0\% IDK and a marked reduction in incorrect responses (from 32.5\% to 10.3\%), signaling a reduction in hallucination frequency.

Although being very effective in reducing hallucination in MCQ datasets, Table \ref{tab:granite_qwen_math} shows that RL-only did not manage to do so in the open-ended QA dataset. This is primarily due to the base model's hesitency to use IDK responses enough that it gets picked up by the RL algorithm which is a classic problem of insufficient exploration in the reinforcement learning domain.

\begin{table}[h!]
\centering
\caption{Comparison of Granite-3.3-2B-Instruct and Qwen-3-4B-Instruct performance on MedMCQA under different reward schemes for RL-only training.}
\resizebox{\columnwidth}{!}{%
\begin{tabular}{llccc}
\toprule
\textbf{Model} & \textbf{IDK Reward} & \textbf{Correct (\%)} & \textbf{Incorrect (\%)} & \textbf{IDK (\%)} \\
\midrule
\multirow{4}{*}{Granite-3.3-2B-Instruct}
& -0.5 & 49.1 & 50.8 & 0.0 \\
& -0.25 & 46.4 & 46.4 & 7.0 \\
& 0 & 15.0 & 6.0 & 78.0 \\
& 0.3 & 0.0 & 0.0 & 100.0 \\
\midrule
\multirow{4}{*}{Qwen-3-4B-Instruct}
& -0.5 & 60.6 & 39.3 & 0.0 \\
& -0.25 & 60.0 & 33.3 & 6.5 \\
& 0 & 55.7 & 27.0 & 16.0 \\
& 0.3 & 48.0 & 10.3 & 41.0 \\
\bottomrule
\end{tabular}%
}
\label{tab:rlonly_rewards_combined}
\end{table}

\subsection{RL-SFT: Teach abstention, then finetune}

In this variant, an additional SFT step is introduced to overcome the model's hesitency to say "I don't know". This is done by training on a dataset with 30\% of randomly selected ground truth set to abstention response. Table~\ref{tab:granite_rlsft_rewards} isolates the effect of reward scaling within RL-SFT training for Granite-3.3-2B-Instruct on the MedMCQA dataset. 
As $r_{abs}$ increases from $-0.5$ to $0.0$, correctness declines monotonically from 39.0\% to 13.0\%, while the IDK rate rises from 3.0\% to 73.0\%. 
This demonstrates a clear trade-off: stronger abstention rewards promote intellectual humility but can suppress the model’s willingness to answer confidently. The optimal balance appears around $r_{abs} = -0.25$, where correctness and abstention remain meaningfully distributed (39.0\% vs. 21.0\%). 
In the Hendrycks Math dataset, as shown in Table \ref{tab:granite_qwen_math}, RL-SFT-Random is able to cut the incorrect answers by half (-28\%) while losing only 10\% in accuracy. 

As seen in Table \ref{tab:granite_qwen_medmcqa}, RL-RTuning underperforms RL-SFT-Random, opposed to what is expected from a method that actually takes into consideration the base models' answers for the SFT step. We believe the major reason for the observed performance gap is the domination of IDK answers in the RL-RTuning after the SFT step ($\sim$ 60\%) compared to RL-SFT-Random ($\sim$ 30\%). The model is unable to recover from this damage during the RL step.

\begin{table}[h!]
\centering
\caption{Comparison of Granite-3.3-2B-Instruct performance on MedMCQA under different Reward IDK values for RL-SFT-Random training.}
\resizebox{\columnwidth}{!}{%
\begin{tabular}{lcccc}
\toprule
\textbf{Reward IDK} & \textbf{Correct (\%)} & \textbf{Incorrect (\%)} & \textbf{IDK (\%)} \\
\midrule
-0.5 & 39.0 & 58.0 & 3.0 \\
-0.4 & 31.0 & 56.0 & 13.0 \\
-0.25 & 39.0 & 40.0 & 21.0 \\
-0.2 & 28.0 & 34.0 & 38.0 \\
-0.1 & 20.0 & 38.0 & 42.0 \\
0.0 & 13.0 & 14.0 & 73.0 \\
\bottomrule
\end{tabular}%
}
\label{tab:granite_rlsft_rewards}
\end{table}

\subsection{RL-RTuning: Teach abstention with RTuning, then finetune}
RL-RTuning is a controlled approach that explicitly teaches the model to abstain (i.e., respond with “I don’t know”) in cases where it lacks sufficient knowledge. Using Granite baselines evaluated on MedMCQA (Table~\ref{tab:granite_rlrtuning_medmcqa_rewards}), we first identify questions that the model answers incorrectly. These incorrect instances are then used as supervision signals to train the model to abstain, while correct instances are reinforced with the correct answers. Compared to RL-SFT with randomized teaching, this method is more proctored and is therefore intuitively expected to yield superior performance. However, as discussed in the previous section, empirical results deviate from this expectation, and we provide justification for this discrepancy.

The same RL-RTuning methodology is applied to the MATH dataset. Table~\ref{tab:granite_rlrtuning_math_rewards} illustrates how varying the abstention reward influences model behavior. As the abstention reward increases, a larger fraction of previously incorrect answers are converted into explicit abstentions. However, this also negatively impacts overall correctness. Consequently, the central challenge is to identify an appropriate abstention reward that balances answer correctness against the model’s willingness to abstain.

\begin{table}[h!]
\centering
\caption{Comparison of Granite-3.3-2B-Instruct performance on MedMCQA under different Reward IDK values for RL-RTuning training.}
\resizebox{\columnwidth}{!}{%
\begin{tabular}{lcccc}
\toprule
\textbf{Reward IDK} & \textbf{Correct (\%)} & \textbf{Incorrect (\%)} & \textbf{IDK (\%)} \\
\midrule
-0.5 & 48.0 & 46.0 & 4.0 \\
-0.25 & 47.0 & 45.0 & 8.0 \\
0.0 & 47.0 & 46.0 & 7.0 \\
0.1 & 46.0 & 49.0 & 5.0 \\
\bottomrule
\end{tabular}%
}
\label{tab:granite_rlrtuning_medmcqa_rewards}
\end{table}

\begin{table}[h!]
\centering
\caption{Comparison of Granite-3.3-2B-Instruct performance on MATH under different Reward IDK values for RL-RTuning training.}
\resizebox{\columnwidth}{!}{%
\begin{tabular}{lcccc}
\toprule
\textbf{Reward IDK} & \textbf{Correct (\%)} & \textbf{Incorrect (\%)} & \textbf{IDK (\%)} \\
\midrule
-0.95 & 25.2 & 69.4 & 5.4 \\
-0.90 & 24.6 & 68.8 & 6.6 \\
-0.80 & 32.4 & 60.0 & 7.6 \\
-0.75 & 29.0 & 58.6 & 12.4 \\
-0.70 & 28.0 & 58.0 & 18.0 \\
-0.60 & 25.2 & 48.8 & 26.0 \\
-0.50 & 21.3 & 33.7 & 44.4 \\
-0.25 & 8.6 & 10.1 & 81.2 \\
-0.10 & 3.4 & 4.3 & 92.3 \\
0.00  & 1.9 & 2.0 & 96.1 \\
0.10  & 0.6 & 1.4 & 98.0 \\
0.25  & 0.7 & 1.1 & 98.2 \\

\bottomrule
\end{tabular}%
}
\label{tab:granite_rlrtuning_math_rewards}
\end{table}

\subsection{Overall Method Comparison}

Table~\ref{tab:granite_qwen_medmcqa} compares Granite-3.3-2B-Instruct and Qwen-3-4B-Instruct across different experimental settings. 
For Granite-3.3-2B-Instruct, adding an IDK option slightly reduces accuracy (46.8\% $\rightarrow$ 44.8\%) while allowing 6.6\% abstentions, indicating early signs of self-uncertainty modeling. 
RL-only training with a moderate abstention penalty ($r_{abs} = -0.25$) achieves a balanced trade-off between correctness (46.4\%) and IDK rate (7.0\%). 
However, further integrating RL with supervised fine-tuning (RL-SFT) increases abstention to 21\%, reflecting stronger uncertainty calibration at some cost to accuracy (39.0\%). 
In contrast, Qwen-3-4B-Instruct maintains higher baseline accuracy (67.5\%) and demonstrates consistent abstention behavior (7.3\%) under the IDK option, while RL-only fine-tuning ($r_{abs} = 0.3$) produces a higher IDK rate (41.0\%) with substantially fewer incorrect answers (10.3\%), suggesting effective hallucination reduction.

Table~\ref{tab:granite_qwen_math} shows that both the RL-SFT variants generalize well to open-ended QA datasets also, where we would expect abstention to be less likely due to the number of possible answers being large, with the Random variant outperforming the RTuning variant yet again. 

\subsection{Metrics}
\begin{figure*}[h!]
\centering
\includegraphics[width=\textwidth]{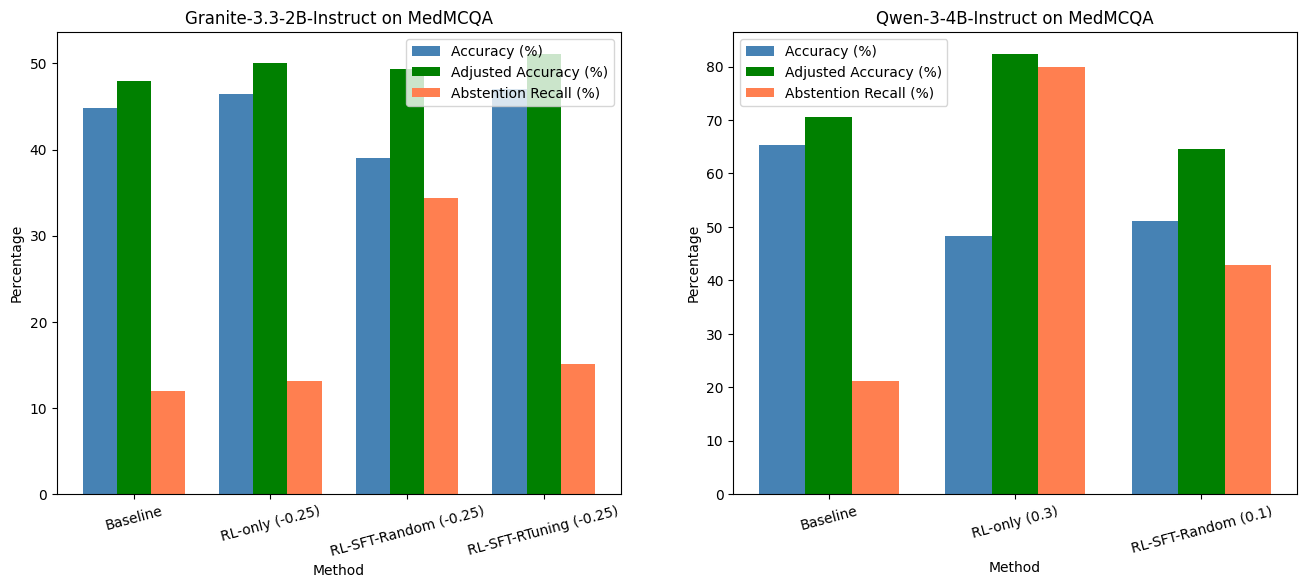}
\caption{Evaluation metrics (accuracy, adjusted accuracy, and abstention recall) for Granite-3.3-2B-Instruct and Qwen-3-2B-Instruct under various training methods on MedMCQA dataset.}
\label{fig:granite_qwen_metrics_medmcqa}
\end{figure*}

\begin{figure}[h!]
\centering
\includegraphics[width=\columnwidth]{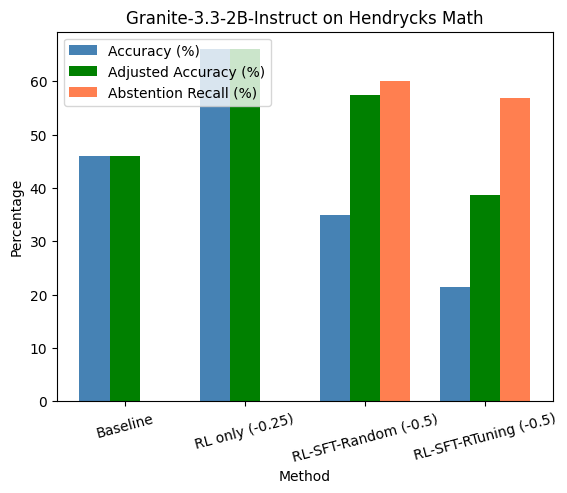}
\caption{Evaluation metrics (accuracy, adjusted accuracy, and abstention recall) for Granite-3.3-2B-Instruct under various training methods on Hendrycks Math.}
\label{fig:granite_metrics_math}
\end{figure}

Figure~\ref{fig:granite_qwen_metrics_medmcqa} and \ref{fig:granite_metrics_math} visualizes Accuracy, Adjusted Accuracy, and Abstention Recall across all methods. In terms of Adjusted Accuracy, RL-only almost always outperforms the other methods but doesn't always do very well in Abstention Recall. Therefore, the results suggest that the most suitable technique highly depends on the scenario and the aspect being optimized for (e.g. hallucination rate, abstention rate, etc...).

\begin{align}
\text{Accuracy} &= \frac{\# Correct}{\# Total\ Samples}\\
\text{Adjusted\ Accuracy} &= \frac{\# Correct}{\# Correct + \# Incorect}\\
\text{Abstention Recall} &= \frac{\# IDK}{\# IDK + \# Incorrect}
\end{align}

\section{Observations}
Across all the different techniques and datasets experimented with, varying the abstention reward $r_{abs}$ systematically modulates model behavior. As $r_{abs}$ increases from negative to positive, abstention rates rise while accuracy declines, revealing a clear trade-off between caution and correctness. Larger models (e.g., Qwen-3-4B-Instruct) maintain higher accuracy under moderate abstention penalties, indicating stronger internal calibration. Qualitatively, RLVR outputs exhibit fewer unsupported claims and more explicit “I don’t know” responses, suggesting reduced hallucination and improved uncertainty awareness. Fine-tuning $r_{abs}$ thus provides a direct control knob over epistemic conservatism—allowing task-specific tuning between confidence and reliability. In addition to that, in models that have a high hesitency to abstain, an initial SFT step helps.

\section{Future Work}

Building upon these promising initial results, our methodology could benefit from exploration in the following directions:

\begin{enumerate}
    \item \textbf{Improved Training Data Mixture for RL-RTuning:} 
    RTuning takes into account the base model's response correctness during the SFT to choose on which questions to abstain. Although our experiments show that the RTuning variant of RL-SFT underperforms, we believe it is mostly due to the SFT step dominated by IDK responses which results in a huge loss in accuracy and general capability of the model in that task. An improvement over our technique could be to control the abstention samples in SFT step to around 20-30\% which will result in a lower probability of model collapse during SFT.
    \item \textbf{Search for a better Random Abstention ratio for RL-SFT-Random:}
    In our experiments, for RL-SFT-Random, we perform SFT with 30\% of randomly chosen responses set to abstention response. Similar to RL-RTuning, we observe a significant loss in acccuracy across both models and both datasets. We believe a better abstention ratio exists, maybe around 10-20\%, which could result in a model that still maintains much of its capabilities after the SFT step.
\end{enumerate}

\section{Conclusion}
We present an abstention-aware extension of Reinforcement Learning with Verifiable Rewards that explicitly incentivizes language models to say “I don’t know” when uncertain. Across multiple models and datasets, we show that tuning the abstention reward provides a direct and effective control over the trade-off between accuracy and hallucination reduction. Our results indicate that moderate abstention rewards can significantly reduce incorrect responses without severely degrading performance, particularly for larger models, while supervised abstention pretraining can improve abstention recall when applied carefully. The findings from experiments on the Hendrycks Math dataset highlight a viable strategy  that, when improved can be a powerful tool in reducing LLM hallucinations even in casual day-to-day conversations.


\bibliography{latex/references}

\clearpage

\onecolumn
\appendix

\section{Qualitative Examples}
In this section, we present a few examples of the model outputs after the abstention training.

\subsection{RL Only - MedMCQA - Granite}
\label{sec:abst_example}
Table~\ref{tab:idk_example_med_granite_rl_only} shows an example where the model correctly abstains on an underspecified MedMCQA-style question after RLVR training.

\begin{table}[h!]
\centering
\small
\begin{tabular}{p{0.95\columnwidth}}
\toprule
\textbf{Question:} What does the vaccine vial monitor indicate? \\[2pt]
\textbf{Options:} \\
A: Colour of outer circle is same as inner square \\
B: Colour of outer circle is darker than inner square \\
C: Colour of outer circle is lighter than inner square \\
D: None of the above \\
E: I Don't Know \\[4pt]
\textbf{Correct Answer:} B \\[4pt]
\textbf{Model Output:} \\
\texttt{<reasoning>} The question lacks sufficient context about the visual semantics of the monitoring device, making a definitive choice uncertain. \texttt{</reasoning>} \\
\texttt{<answer>E</answer>} \\
\bottomrule
\end{tabular}
\caption{Example illustrating abstention behavior. The model selects the explicit ``I Don’t Know'' option instead of guessing, demonstrating reduced hallucination.}
\label{tab:idk_example_med_granite_rl_only}
\end{table}

\subsection{RL-SFT-Random - MedMCQA - Granite}
\label{sec:abst_example_med_granite_rlsft}

Table~\ref{tab:idk_example_med_granite_rlsft} shows an example where the model abstains despite partial domain knowledge.

\begin{table}[h!]
\centering
\small
\begin{tabular}{p{0.95\columnwidth}}
\toprule
\textbf{Question:} All the following familial syndromes are associated with development of pheochromocytomas except? \\[2pt]
\textbf{Options:} \\
A: Sturge-Weber syndrome \\
B: Von Recklinghausen disease \\
C: MEN Type II \\
D: Prader-Willi syndrome \\
E: I Don't Know \\[4pt]
\textbf{Correct Answer:} D \\[4pt]
\textbf{Model Output:} \\
\texttt{<reasoning>} Pheochromocytomas are seen in MEN II and von Recklinghausen disease. I am not confident in my reasoning and choose to abstain. \texttt{</reasoning>} \\
\texttt{<answer>E</answer>} \\
\bottomrule
\end{tabular}
\caption{RL-SFT-Random MedMCQA example where the model abstains despite partial medical knowledge.}
\label{tab:idk_example_med_granite_rlsft}
\end{table}

\subsection{RL-SFT-Random - Hendrycks Math - Granite}
Here's a example where the model, although seems to solve the problem with proper proof, ends up with a wrong answer. But it chooses to abstain from answering since it doesn't feel entirely confident, which turns out to be the right choice in this situation due to the correct answer being entirely different.

\textbf{Question:}
\begin{quote}
Let $a,$ $b,$ $c$ be nonnegative real numbers such that $a + b + c = 1.$  Find the maximum value of \[\frac{ab}{a + b} + \frac{ac}{a + c} + \frac{bc}{b + c}\]
\end{quote}

\textbf{Correct Answer:} $\frac{1}{2}$

\textbf{Model Response:}
\begin{quote}
By AM-HM,
\begin{align*}
\frac{a + b}{2} &\ge \frac{2}{\frac{1}{a} + \frac{1}{b}}, \\
\frac{a + c}{2} &\ge \frac{2}{\frac{1}{a} + \frac{1}{c}}, \\
\frac{b + c}{2} &\ge \frac{2}{\frac{1}{b} + \frac{1}{c}}.
\end{align*}

Then
\begin{align*}
\frac{ab}{a + b} &\le \frac{a}{2} \cdot \frac{a}{a + b}, \\
\frac{ac}{a + c} &\le \frac{a}{2} \cdot \frac{a}{a + c}, \\
\frac{bc}{b + c} &\le \frac{b}{2} \cdot \frac{b}{b + c}.
\end{align*}

Adding all these inequalities, we get
\begin{align*}
&\frac{ab}{a + b} + \frac{ac}{a + c} + \frac{bc}{b + c} \le \\&\frac{a}{2} \cdot \left( \frac{a}{a + b} + \frac{a}{a + c} + \frac{a}{b + c} \right) + \frac{b}{2} \cdot \left( \frac{b}{a + b} + \frac{b}{a + c} + \frac{b}{b + c} \right) + \frac{c}{2} \cdot \left( \frac{c}{a + b} + \frac{c}{a + c} + \frac{c}{b + c} \right).
\end{align*}

By AM-HM again,
\begin{align*}
\frac{a}{a + b} + \frac{a}{a + c} + \frac{a}{b + c} &\ge \frac{3a^2}{2(a + b) + 2(a + c) + 2(b + c)} = \frac{3a^2}{4a + 4b + 4c} = \frac{3a^2}{4(a + b + c)} = \frac{3a^2}{4}, \\
\frac{b}{a + b} + \frac{b}{a + c} + \frac{b}{b + c} &\ge \frac{3b^2}{4(a + b + c)} = \frac{3b^2}{4}, \\
\frac{c}{a + b} + \frac{c}{a + c} + \frac{c}{b + c} &\ge \frac{3c^2}{4(a + b + c)} = \frac{3c^2}{4}.
\end{align*}

Hence,
\begin{align*}
\frac{ab}{a + b} + \frac{ac}{a + c} + \frac{bc}{b + c} &\le \frac{a}{2} \cdot \frac{3a}{4} + \frac{b}{2} \cdot \frac{3b}{4} + \frac{c}{2} \cdot \frac{3c}{4} \\
&= \frac{3}{8} (a^2 + b^2 + c^2) \\
&\le \frac{3}{8} (a + b + c)^2 \\
&= \frac{3}{8},
\end{align*}

with equality when $a = b = c = \frac{1}{3}.$  Therefore, the maximum value is $\boxed{\frac{3}{8}}.$ I'm unsure whether my reasoning holds up, so I'll avoid answering.\boxed{I\ Don't\ Know}
\end{quote}

\subsection{RL-RTuning - Hendrycks Math - Granite}
Here's another example where the model, seems to solve the problem but it does not count the number of lattice points correctly. $(\pm 2, \pm 2, \pm 1)$ should be counted with 24 lattice points. It chooses to not give answer as it is uncertain.

\textbf{Question:}
\begin{quote}
In three-dimensional space, find the number of lattice points that have a distance of 3 from the origin.

Note: A point is a lattice point if all its coordinates are integers.
\end{quote}

\textbf{Correct Answer:} 30

\textbf{Model Response:}
\begin{quote}
<reasoning>Let the three-dimensional space have a vertex at the origin $(0,0,0)$, and let a lattice point be any point $(x,y,z)$ with integer coordinates.  Then by the distance formula, \[x^2 + y^2 + z^2 = 3^2 = 9.\]It follows that $x^2, y^2, z^2 \le 9$, and that $x, y, z \in \{-3, -2, -1, 0, 1, 2, 3\}$.

We can find the six points $(\pm 3, 0, 0)$, $(0, \pm 3, 0)$, and $(0, 0, \pm 3)$.  We can also find $(0, \pm 1, \pm 2)$ and $(\pm 1, \pm 2, 0)$ by checking all possible combinations.  This gives us $\boxed{27}$ lattice points. I'm uncertain about my conclusions, so I'd prefer not to answer.</reasoning><answer>\boxed{\text{IDK}}</answer>
\end{quote}

\end{document}